\documentclass[11pt]{article}

\usepackage[T1]{fontenc}
\usepackage[margin=1in]{geometry}
\usepackage{lmodern}
\usepackage{microtype}
\usepackage{amsmath,amssymb,amsfonts,amsthm}
\usepackage{mathtools}
\usepackage{booktabs}
\usepackage{graphicx}
\usepackage{hyperref}
\usepackage{cleveref}
\usepackage{enumitem}
\usepackage{placeins}
\usepackage{natbib}
\setcitestyle{authoryear,open={(},close={)}}

\newcommand{\R}{\mathbb{R}}
\newcommand{\C}{\mathbb{C}}
\newcommand{\U}{\mathrm{U}}
\newcommand{\SU}{\mathrm{SU}}
\newcommand{\SO}{\mathrm{SO}}
\newcommand{\Orth}{\mathrm{O}}
\newcommand{\tr}{\operatorname{tr}}
\newcommand{\diag}{\operatorname{diag}}
\renewcommand{\skew}{\mathrm{skew}}
\DeclareMathOperator{\Exp}{exp}

\theoremstyle{definition}

\title{Subgroups of $\U(d)$ Induce Natural RNN and Transformer Architectures}
\author{Joshua Nunley\\
Cognitive Science Program \& Department of Informatics\\
Indiana University\\
\texttt{joshnunl@iu.edu}}
\date{February 20, 2026}

\begin{document}
\maketitle

\begin{abstract}
This paper presents a direct framework for sequence models with hidden states on closed subgroups of $\U(d)$. We use a minimal axiomatic setup and derive recurrent and transformer templates from a shared skeleton in which subgroup choice acts as a drop-in replacement for state space, tangent projection, and update map. We then specialize to $\Orth(d)$ and evaluate orthogonal-state RNN and transformer models on Tiny Shakespeare and Penn Treebank under parameter-matched settings. We also report a general linear-mixing extension in tangent space, which applies across subgroup choices and improves finite-budget performance in the current $\Orth(d)$ experiments.
\end{abstract}

\section{Introduction}

Modern sequence models are built from many interacting architectural choices: update parameterization, stabilization layers, memory mechanisms, and readout conventions. This paper takes a narrower starting point. We assume hidden states and token representations are group-valued, specifically in a closed subgroup of $\U(d)$, and ask what recurrent and transformer architectures follow directly from that assumption.

The motivation for this choice is methodological. A compact matrix group gives bounded state dynamics by construction, a Lie algebra for local updates, and a natural similarity/readout based on matrix inner products. These ingredients are enough to define complete models without introducing many additional components.

This manuscript has three concrete goals. First, we give a minimal framework and model template based on closed subgroups of $\U(d)$. Second, we show how subgroup choice can be used as a drop-in model design axis for both RNN and transformer templates. Third, we report experiments for one subgroup family, $\Orth(d)$, using the resulting orthogonal-state RNN and transformer instantiations.

\section{Related Work and Positioning}

\paragraph{Orthogonal and unitary RNNs.}
A long line of work stabilizes recurrent dynamics by constraining the transition operator to be orthogonal or unitary \citep{arjovsky2016unitary,wisdom2016full,helfrich2018orthogonal,lezcano2019cheap}, with recent variants exploring different parameterizations and update rules \citep{biegun2024rotrnn,alkhairy2025deltaproduct}. In all of these models, hidden states remain Euclidean vectors; the group structure constrains the operator, not the state itself.

\paragraph{Manifold-valued states and group-aware attention.}
The closest prior direction to ours is SPD-based sequence modeling \citep{seraphim2024spd,wang2023smsa,dubreil2024riemannian}, where hidden states live on a matrix manifold rather than in $\R^n$. However, those models use symmetric positive-definite matrices (a symmetric space), not compact group elements with Lie-algebraic tangent updates. Separately, group-aware attention constructions \citep{hutchinson2021lie,fuchs2020se3} and orthogonality constraints on attention operators \citep{zhang2026osa} use group structure for equivariant processing of inputs, not for the hidden-state dynamics themselves.

\paragraph{This work.}
Our contribution is twofold. First, we place both hidden states and token representations directly on a closed subgroup of $\U(d)$, rather than keeping states in $\R^n$ and constraining operators. Second, we give a shared RNN/transformer template in which subgroup choice is a direct drop-in component---changing the state space, tangent projection, and update map without altering the outer model interface. To our knowledge, neither of these choices has appeared in this combined form for both recurrent and transformer models. We evaluate one instantiation ($\Orth(d)$) under parameter-matched settings.

\section{Framework and Axioms}

Let $G \leq \U(d)$ be a closed Lie subgroup with Lie algebra $\mathfrak g$. We represent each token $x_t$ by a subgroup element $M_{x_t} \in G$ and maintain state $H_t \in G$.

We use the following minimal assumptions.

\begin{enumerate}[leftmargin=1.5em]
  \item \textbf{Causality:} $H_{t+1}$ depends only on $(H_t, x_t)$.
  \item \textbf{Group closure:} if $H_t \in G$, then $H_{t+1} \in G$.
  \item \textbf{Local update:} each step is generated by a tangent element in $\mathfrak g$.
\end{enumerate}

A standard realization is a multiplicative step generated by the exponential map. Concretely, for any measurable map $\Phi_\theta(H_t,x_t)\in\mathfrak g$,
\[
H_{t+1} = H_t\,\Exp\!\big(\Phi_\theta(H_t,x_t)\big)
\]
remains in $G$ for all $t$ by subgroup closure under $\Exp(\mathfrak g)$ and group multiplication. We treat this as the construction rule rather than a standalone theorem.

This gives a single model skeleton across subgroup choices: choose $G$, define $\Phi_\theta$ in $\mathfrak g$, and read out predictions from $H_t$.

For readout, we use prototype matrices $P_v \in G$ and logits
\[
\ell_v(H_t) = \tau\,\Re\,\tr(H_t^\ast P_v) + b_v.
\]
This is the default subgroup-native readout used throughout this paper.

From a Riemannian viewpoint, this is the natural local-geometry construction: updates are in tangent space ($\mathfrak g$), then mapped back to the manifold by $\Exp$.

\section{General RNN and Transformer Templates}

This section defines subgroup-agnostic templates. The subgroup enters only through the state space $G$, tangent space $\mathfrak g$, and projection/update operators.

\subsection{RNN template}
Given a sequence $(x_t)_{t=1}^T$, initialize $H_0 \in G$ and update
\[
H_{t+1}=H_t\,\Exp\big(U_t\big), \qquad U_t\in\mathfrak g.
\]
We first construct a raw tangent update
\[
\widehat U_t=\Phi_\theta(H_t,M_{x_t})\in\mathfrak g,
\]
then optionally apply a tangent-space map
\[
U_t=\Gamma_\theta(\widehat U_t), \qquad \Gamma_\theta:\mathfrak g\to\mathfrak g.
\]
Token logits are computed from $H_t$ with the subgroup-native prototype head
\[
\ell_v(H_t) = \tau\,\Re\,\tr(H_t^\ast P_v) + b_v.
\]

\subsection{Transformer template}
For each position $i$, initialize $H_i^{(0)}\in G$ from token embeddings. At each layer $\ell$, compute causal attention weights from subgroup similarity scores,
\[
s_{ij}^{(\ell)} = \tau\,\Re\,\tr\!\left((H_i^{(\ell)})^\ast H_j^{(\ell)}\right),
\qquad j\le i,
\]
\[
\alpha_{ij}^{(\ell)} = \mathrm{softmax}_j\big(s_{ij}^{(\ell)} + b_{ij}^{(\ell)}\big),
\]
then apply two subgroup-tangent sub-steps. The \emph{attention step} updates the state toward the attention-weighted aggregate:
\[
\widehat\Delta_{i,\mathrm{attn}}^{(\ell)} =
\Psi_{\theta,\mathrm{attn}}\!\left(H_i^{(\ell)},\sum_{j\le i}\alpha_{ij}^{(\ell)}H_j^{(\ell)}\right)\in\mathfrak g,
\]
\[
\Delta_{i,\mathrm{attn}}^{(\ell)}=
\Gamma_{\theta,\mathrm{attn}}\!\big(\widehat\Delta_{i,\mathrm{attn}}^{(\ell)}\big),\qquad
\widetilde H_i^{(\ell)} = H_i^{(\ell)}\,\Exp\big(\Delta_{i,\mathrm{attn}}^{(\ell)}\big).
\]
The \emph{grounding step} then adjusts toward the current token embedding:
\[
\widehat\Delta_{i,\mathrm{ground}}^{(\ell)} =
\Psi_{\theta,\mathrm{ground}}\!\left(\widetilde H_i^{(\ell)},M_{x_i}\right)\in\mathfrak g,
\]
\[
\Delta_{i,\mathrm{ground}}^{(\ell)}=
\Gamma_{\theta,\mathrm{ground}}\!\big(\widehat\Delta_{i,\mathrm{ground}}^{(\ell)}\big),\qquad
H_i^{(\ell+1)} = \widetilde H_i^{(\ell)}\,\Exp\big(\Delta_{i,\mathrm{ground}}^{(\ell)}\big).
\]
In this template we do not introduce separate learned query/key/value maps. The score
\(
\Re\,\tr((H_i^{(\ell)})^\ast H_j^{(\ell)})
\)
is already a coordinate-invariant subgroup similarity under global basis changes (conjugation), so attention can be computed directly from the group-valued states.
Final logits come from $H_i^{(L)}$ via the same prototype head as in the RNN template.

\subsection{Optional tangent mixing component}
Both templates include the same optional tangent map component
\[
\Gamma:\mathfrak g\to\mathfrak g
\]
applied between raw tangent construction and the exponential group step. The intrinsic base case is $\Gamma=\mathrm{Id}$. The symmetry-relaxed variant uses a learned linear map in tangent coordinates:
\[
a=\mathrm{vec}_{\mathfrak g}(A),\qquad a'=Wa,\qquad \Gamma(A)=\mathrm{vec}_{\mathfrak g}^{-1}(a').
\]
This applies to recurrent updates and to both transformer sub-steps (attention and grounding).

With USIM readout, $\Gamma$ has a concrete channel interpretation. At any state $H$, the readout differential
\[
D_H(A)=\left.\tfrac{d}{d\epsilon}\right|_{\epsilon=0}\ell(H\Exp(\epsilon A))
\]
induces a splitting of the tangent space into readout-visible and readout-null directions:
\[
\mathfrak g=\underbrace{\ker(D_H)^\perp}_{\text{predictive}}\;\oplus\;\underbrace{\ker(D_H)}_{\text{memory}}.
\]
In tangent coordinates, $\Gamma$ can therefore reweight and route updates across these two channels before exponentiation. The identity map preserves the raw geometry; a learned linear map can reallocate capacity between channels.

These equations define the templates used in the rest of the paper. Subgroup choice changes the internals of $\Phi_\theta$ and $\Psi_\theta$, not the outer training interface.

\section{Subgroup Instantiations as Drop-In Components}

The framework applies to any closed Lie subgroup of $\U(d)$. In this paper we focus on the following families: $\U(d)$, $\SU(d)$, $\Orth(d)$, $\SO(d)$, and torus models $T^k$. Each subgroup yields a concrete drop-in specification of tangent projection and state update. This section stays at component level; the exact $\Orth(d)$ equations used in runs are written at the start of the experiments section. Full parameterization recipes for all subgroup families (embedding construction, tangent-map choices, and update-map approximations) are collected in \cref{sec:components}.

\begin{table}[ht]
\centering
\small
\begin{tabular}{l p{2.0cm} p{3.0cm} p{3.0cm}}
\toprule
Subgroup & Tangent space & Tangent projection & Practical effect \\
\midrule
$\U(d)$ & $\mathfrak u(d)$ & $A\leftarrow\tfrac12(A-A^\ast)$ & full unitary updates \\
$\SU(d)$ & $\mathfrak{su}(d)$ & skew-Hermitian with trace removal & phase-free unitary updates \\
$\Orth(d)$/$\SO(d)$ & $\mathfrak{so}(d)$ & $A\leftarrow\tfrac12(A-A^\top)$ & real orthogonal updates \\
$T^k$ & diagonal imaginary algebra & keep diagonal tangent terms & abelian channel dynamics \\
\bottomrule
\end{tabular}
\caption{Subgroup-specific drop-in components for the shared RNN/transformer templates.}
\label{tab:subgroup_dropins}
\end{table}

In all cases, the state update keeps the same external form
\[
H \leftarrow H\,\Exp(A), \qquad A\in\mathfrak g,
\]
while the subgroup-specific projection defines how raw model outputs are mapped into $\mathfrak g$.

The concrete choice used here is to build tangent updates from relative state terms, then project:
\[
\Psi_{\mathrm{attn}}(H,V)=\Gamma_{\mathrm{attn}}\!\big(\Pi_{\mathfrak g}(H^\ast V)\big),
\qquad
\Psi_{\mathrm{ground}}(H,M)=\Gamma_{\mathrm{ground}}\!\big(\Pi_{\mathfrak g}(H^\ast M)\big),
\]
where $\Pi_{\mathfrak g}$ is the subgroup-specific tangent projection (table row), and $\Gamma_{\mathrm{attn}},\Gamma_{\mathrm{ground}}:\mathfrak g\to\mathfrak g$ are the template-level tangent maps from the general templates.

The default readout across all subgroup choices is the untied similarity head
\[
\ell_v(H)=\tau\,\Re\,\tr(H^\ast P_v)+b_v,\qquad P_v\in G.
\]

\subsection{Scope of experiments}
The experiments evaluate $\Orth(d)$ models only. The $\Orth(d)$ equations are obtained by direct substitution:
\[
\Pi_{\mathfrak g}(Y)=\tfrac12(Y-Y^\top),\qquad G=\Orth(d),\qquad \mathfrak g=\mathfrak{so}(d).
\]
The experiments section writes out the full evaluated equations with the approximation choices used in reported runs.

\section{Experiments on \texorpdfstring{$\Orth(d)$}{O(d)} Models}

We evaluate character-level language modeling on Tiny Shakespeare (TS) and Penn Treebank (PTB), reporting bits per character (BPC; lower is better). Unless stated otherwise, models are trained with Adam (learning rate $10^{-3}$, weight decay $10^{-4}$), batch size 32, gradient clipping at 1.0, and early stopping patience of 50 epochs. Results reported here are single-seed.

All OSM results in the main tables use the subgroup-native USIM readout; linear Euclidean readout appears only in ablations.

\subsection{Evaluated equations and chosen approximations}
For reported $\Orth(d)$ runs, we use
\[
\Pi_{\mathfrak g}(Y)=\skew(Y)=\tfrac12(Y-Y^\top),\qquad
H\leftarrow H\,\Exp(A).
\]
Token embeddings and USIM prototypes are parameterized on-group as
\[
M_v=\Exp(\skew(B_v)),\qquad P_v=\Exp(\skew(C_v)).
\]
For OSMFormer, layer updates are
\[
Z_i^{(\ell)}=\Gamma_{\mathrm{attn}}\!\left(
\skew\!\left((H_i^{(\ell)})^\top\sum_{j\le i}\alpha_{ij}^{(\ell)}H_j^{(\ell)}\right)\right),
\]
\[
G_i^{(\ell)}=\Gamma_{\mathrm{ground}}\!\left(
\skew\!\left((\widetilde H_i^{(\ell)})^\top M_{x_i}\right)\right),
\]
\[
\widetilde H_i^{(\ell)} = H_i^{(\ell)}\Exp(Z_i^{(\ell)}),\qquad
H_i^{(\ell+1)} = \widetilde H_i^{(\ell)}\Exp(G_i^{(\ell)}),
\]
where $\Gamma_{\mathrm{attn}},\Gamma_{\mathrm{ground}}$ are either identity (no mixing) or learned linear maps in skew-coordinate space (linear mixing), as defined in the general templates. OSM-RNN uses the same pattern with its recurrent tangent update.

\subsection{Parameter-matched comparisons}
Tables~\ref{tab:ts_100k_main}--\ref{tab:ts_rnn_52k_main} compare OSM models against standard baselines \citep{vaswani2017attention,press2022alibi} at fixed parameter budgets. Both the transformer and RNN variants use linear tangent mixing.

\begin{table}[ht]
\centering
\caption{Tiny Shakespeare, approximately 100K parameters (single-seed; OSM row uses USIM readout).}
\label{tab:ts_100k_main}
\begin{tabular}{lrrr}
\toprule
Model & Params & Val BPC & Test BPC \\
\midrule
Transformer (ALiBi, 2L1H) & 104{,}111 & 2.259 & 2.583 \\
OSMFormer (USIM, ALiBi, linear mixing) & 91{,}429 & \textbf{2.254} & \textbf{2.464} \\
\bottomrule
\end{tabular}
\end{table}

\begin{table}[ht]
\centering
\caption{Penn Treebank, approximately 100K parameters (single-seed; OSM row uses USIM readout).}
\label{tab:ptb_100k_main}
\begin{tabular}{lrrr}
\toprule
Model & Params & Val BPC & Test BPC \\
\midrule
Transformer (ALiBi, 2L1H) & 101{,}816 & 1.716 & 1.664 \\
OSMFormer (USIM, ALiBi, linear mixing) & 103{,}482 & \textbf{1.665} & \textbf{1.614} \\
\bottomrule
\end{tabular}
\end{table}

\begin{table}[ht]
\centering
\caption{Tiny Shakespeare, approximately 52K parameters (single-seed RNN comparison; OSM row uses USIM readout).}
\label{tab:ts_rnn_52k_main}
\begin{tabular}{lrrr}
\toprule
Model & Params & Val BPC & Test BPC \\
\midrule
LSTM & 52{,}305 & 2.407 & 2.594 \\
OSM-RNN (USIM, linear mixing) & 55{,}545 & \textbf{2.366} & \textbf{2.539} \\
\bottomrule
\end{tabular}
\end{table}

\subsection{Scaling summaries}
We compare OSMFormer (USIM, linear mixing) against the baseline transformer across parameter budgets from 100K to 500K on both datasets.

\begin{table}[ht]
\centering
\caption{Tiny Shakespeare scaling (single-seed): baseline transformer vs OSMFormer (USIM, linear mixing).}
\label{tab:ts_scaling_summary}
\begin{tabular}{lrrrr}
\toprule
Budget & Baseline Params & Baseline Val/Test & OSM Params & OSM Val/Test \\
\midrule
100K & 104{,}111 & 2.259 / 2.583 & 91{,}429 & \textbf{2.254} / \textbf{2.464} \\
300K & 299{,}047 & 2.207 / 2.552 & 277{,}357 & \textbf{2.184} / \textbf{2.377} \\
400K & 392{,}987 & 2.210 / 2.559 & 380{,}757 & \textbf{2.184} / \textbf{2.373} \\
500K & 499{,}727 & 2.229 / 2.493 & 511{,}749 & \textbf{2.176} / \textbf{2.365} \\
\bottomrule
\end{tabular}
\end{table}

\begin{table}[ht]
\centering
\caption{Penn Treebank scaling (single-seed): baseline transformer vs OSMFormer (USIM, linear mixing).}
\label{tab:ptb_scaling_summary}
\begin{tabular}{lrrrr}
\toprule
Budget & Baseline Params & Baseline Val/Test & OSM Params & OSM Val/Test \\
\midrule
300K & 295{,}072 & 1.596 / 1.551 & 310{,}002 & \textbf{1.570} / \textbf{1.524} \\
400K & 408{,}616 & 1.568 / 1.523 & 423{,}754 & \textbf{1.557} / \textbf{1.512} \\
500K & 494{,}552 & 1.558 / 1.512 & 491{,}454 & \textbf{1.549} / \textbf{1.503} \\
\bottomrule
\end{tabular}
\end{table}

\begin{figure}[t]
\centering
\includegraphics[width=0.9\linewidth]{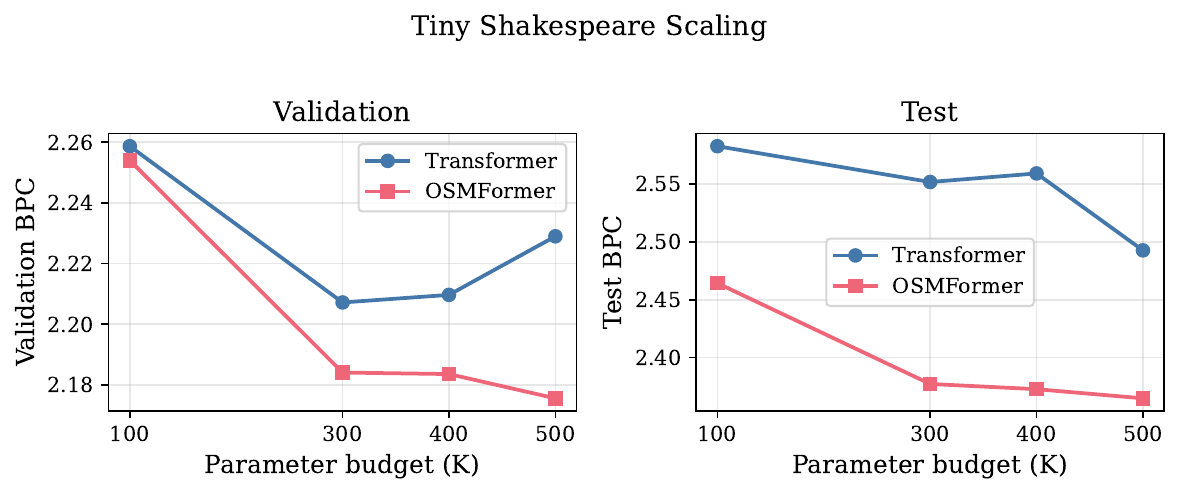}
\caption{Tiny Shakespeare scaling curve corresponding to Table~\ref{tab:ts_scaling_summary}.}
\label{fig:ts_scaling_curve}
\end{figure}

\begin{figure}[t]
\centering
\includegraphics[width=0.9\linewidth]{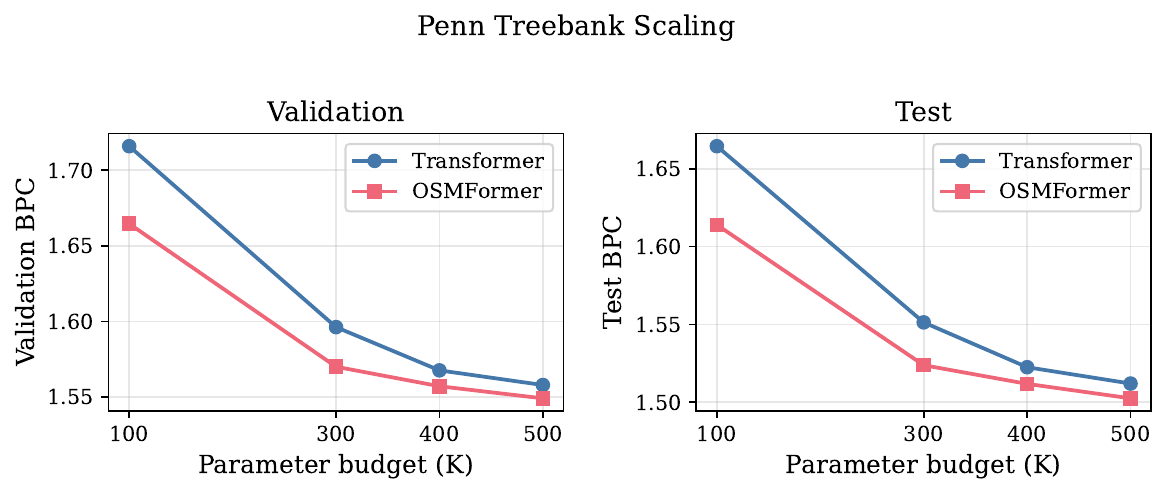}
\caption{Penn Treebank scaling curve corresponding to Table~\ref{tab:ptb_scaling_summary}.}
\label{fig:ptb_scaling_curve}
\end{figure}

\subsection{Linear mixing and robustness summaries}

Among tested update variants, linear mixing is the best-performing option in this $\Orth(d)$ setting.
In the table below, ``BCH-style intrinsic mix'' denotes a second-order Lie-bracket correction update used instead of the learned linear tangent map.

\begin{table}[ht]
\centering
\caption{Tiny Shakespeare, 100K OSMFormer mixing ablation (single-seed, USIM readout fixed across rows).}
\label{tab:mix_ablation_main}
\begin{tabular}{lrrr}
\toprule
Mixing mode & Params & Val BPC & Test BPC \\
\midrule
No tangent mixing & 103{,}501 & 2.355 & 2.525 \\
BCH-style intrinsic mix & 103{,}503 & 2.349 & 2.532 \\
Linear tangent mixing & 91{,}429 & \textbf{2.254} & \textbf{2.464} \\
\bottomrule
\end{tabular}
\end{table}

The linear-mixing variant has fewer parameters (91K vs 103K) due to different hidden dimensions in the matched configurations, yet still attains the best BPC.

For optimizer robustness, we also include the 500K Tiny Shakespeare sweep across 9 settings. OSMFormer remains trainable across all tested settings and has a narrower best-validation range than the baseline under the same sweep.

\begin{figure}[t]
\centering
\includegraphics[width=0.9\linewidth]{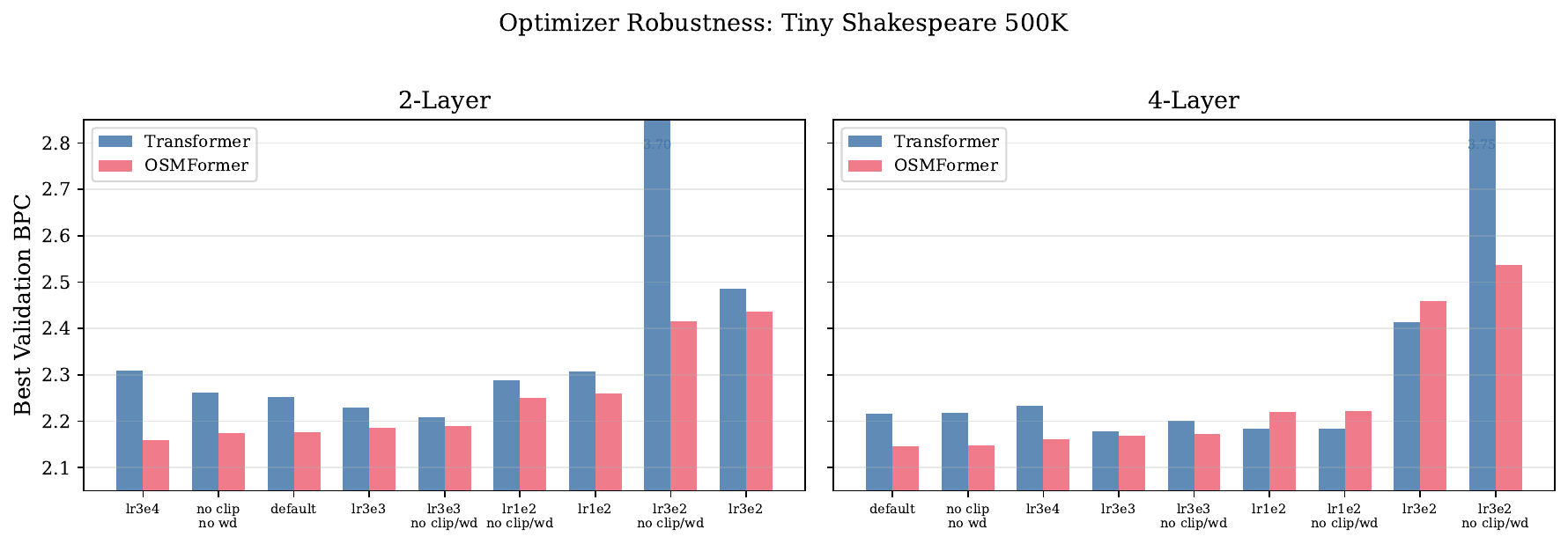}
\caption{Optimizer robustness summary on Tiny Shakespeare at 500K parameters (2-layer and 4-layer settings).}
\label{fig:ts_robustness_curve}
\end{figure}

\FloatBarrier

\section{Limitations and Scope}

Experiments focus on character-level Tiny Shakespeare and Penn Treebank and are reported as single-seed runs. The framework itself is subgroup-generic, but this manuscript only validates the $\Orth(d)$ instantiation in depth. Wider empirical coverage across subgroup families (for example $\SU(d)$ and $T^k$) is left to follow-up work.

\section{Discussion and Future Work}

The current results establish that the subgroup-template construction is usable in practice for the $\Orth(d)$ case. The next stage is to evaluate what this structure contributes beyond held-out accuracy metrics. We highlight three concrete directions.

\paragraph{Interpretability.}
The model family defines updates in tangent space and then maps them back to the group through $\Exp$, which creates a direct handle on update geometry. A practical next step is to track per-step tangent magnitudes, per-plane update activity, and attention-vs-grounding contributions during training and inference. On the theory side, these diagnostics can be connected to local linearization of readout and to subgroup-specific coordinate choices for comparing runs across seeds. On the empirical side, the goal is to test whether these quantities are stable and predictive of failure modes, not only descriptive after the fact.

\paragraph{Stability.}
Compact-state dynamics suggest strong training stability, but this should be quantified more systematically. We plan multi-seed robustness sweeps, longer training windows, and broader optimizer regimes (including more aggressive learning-rate and depth settings) under fixed parameter budgets. A parallel theoretical direction is to formalize finite-step sensitivity bounds under subgroup updates and to characterize how stability changes under symmetry-relaxed options such as linear mixing.

\paragraph{Memory.}
The subgroup view naturally induces a memory-design axis: abelian and near-abelian subgroups versus fully non-abelian subgroups imply different interaction patterns in tangent updates. A direct experimental program is to compare these families on tasks with controlled long-range dependencies while holding training protocol and budget constant. A complementary theoretical program is to relate subgroup algebraic structure and update parameterization to measurable memory behavior (retention, transfer, and horizon).

These directions are intentionally incremental: they can be executed within the same template and differ mainly by subgroup drop-ins, readout/mixing choices, and evaluation protocol.

\section{Conclusion}

This paper presented a direct sequence-model framework built from closed subgroups of $\U(d)$. We gave shared RNN and transformer templates, showed how subgroup choice acts as a drop-in component set, and instantiated the framework concretely on $\Orth(d)$.

Under parameter-matched settings on Tiny Shakespeare and Penn Treebank, the resulting orthogonal-state RNN and transformer models are competitive with standard baselines. A linear tangent-mixing extension, which applies to any subgroup in the framework, provides the strongest finite-budget performance in the current $\Orth(d)$ experiments.

The main output of this manuscript is a straightforward construction pipeline: choose a subgroup, plug its tangent projection and update map into the shared templates, and train with the same autoregressive objective. Further subgroup comparisons and larger-scale evaluations are natural next steps.

\section*{Acknowledgments}

\paragraph{Declaration of generative AI use.}
Generative AI tools were used to assist with manuscript drafting,
literature review, and code editing. All content was reviewed,
verified, and revised by the author, who takes full responsibility for
the accuracy and integrity of the work.

\paragraph{Compute resources.}
The authors acknowledge the Indiana University Pervasive Technology Institute for providing supercomputing and storage resources that contributed to this work, including Big Red 200. This research was supported in part by Lilly Endowment, Inc., through its support for the Indiana University Pervasive Technology Institute.

\bibliography{references}

\appendix
\section{Component Approximations and Parameterizations}
\label{sec:components}

The main text defines the templates and subgroup drop-in components in exact form.
This section specifies implementation choices for all subgroup families.

\subsection{On-group embedding and readout parameterization}
For each token $v$, we keep two trainable parameters:
\begin{itemize}[leftmargin=1.5em]
  \item $B_v$: input-embedding parameter (used to build $M_v$),
  \item $C_v$: readout-prototype parameter (used to build $P_v$).
\end{itemize}
For matrix-group cases, we first project to the Lie algebra and then exponentiate:
\[
A_v = \Pi_{\mathfrak g}(B_v),\qquad
Q_v = \Pi_{\mathfrak g}(C_v),\qquad
M_v=\Exp(A_v),\qquad P_v=\Exp(Q_v).
\]
So $B_v,C_v$ are unconstrained parameters, while $M_v,P_v\in G$ are the actual on-group objects used by the model.
The same construction is used for input embeddings ($M_v$) and readout prototypes ($P_v$).

\paragraph{Compact notation.}
To keep formulas readable, we define
\[
\skew_{\C}(X):=\tfrac12(X-X^\ast),\qquad
\skew_{\R}(X):=\tfrac12(X-X^\top),\qquad
\mathrm{tr0}(X):=X-\tfrac1d\tr(X)I.
\]

\begin{table}[ht]
\centering
\small
\begin{tabular}{llll}
\toprule
Subgroup & Raw parameters & Projection to $\mathfrak g$ & On-group map \\
\midrule
$\U(d)$
& $B_v,C_v\in\C^{d\times d}$
& $\skew_{\C}$
& $\Exp$ \\

$\SU(d)$
& $B_v,C_v\in\C^{d\times d}$
& $\mathrm{tr0}\circ\skew_{\C}$
& $\Exp$ \\

$\SO(d)$
& $B_v,C_v\in\R^{d\times d}$
& $\skew_{\R}$
& $\Exp$ \\

$\Orth(d)$
& $B_v,C_v\in\R^{d\times d}$
& $\skew_{\R}$
& $F^{\eta_v}\Exp(\cdot)$ for $M_v$, $F^{\xi_v}\Exp(\cdot)$ for $P_v$ \\

$T^k$
& $\beta_v,\gamma_v\in\R^k$
& $i\,\diag(\cdot)$
& $\Exp$ (equivalently elementwise complex phase) \\

\bottomrule
\end{tabular}
\caption{Embedding/prototype parameterization by subgroup (operator form). For $\Orth(d)$, $F=\diag(-1,1,\dots,1)$ and $\eta_v,\xi_v\in\{0,1\}$. Exact token-wise formulas are given in the recipe list below.}
\label{tab:subgroup_approx_components}
\end{table}

\paragraph{Concrete recipes used in code.}
For clarity, the token construction for each subgroup is:
\begin{itemize}[leftmargin=1.5em]
  \item $\U(d)$: $M_v=\Exp(\skew_{\C}(B_v))$, $P_v=\Exp(\skew_{\C}(C_v))$.
  \item $\SU(d)$: $M_v=\Exp(\mathrm{tr0}(\skew_{\C}(B_v)))$, $P_v=\Exp(\mathrm{tr0}(\skew_{\C}(C_v)))$.
  \item $\SO(d)$: $M_v=\Exp(\skew_{\R}(B_v))$, $P_v=\Exp(\skew_{\R}(C_v))$.
  \item $\Orth(d)$: with $F=\diag(-1,1,\dots,1)$ and $\eta_v,\xi_v\in\{0,1\}$,
  \[
  M_v=F^{\eta_v}\Exp(\skew_{\R}(B_v)),\qquad
  P_v=F^{\xi_v}\Exp(\skew_{\R}(C_v)).
  \]
  Reported runs use $\eta_v=\xi_v=0$.
  \item $T^k$: $M_v=\Exp(i\,\diag(\beta_v))$, $P_v=\Exp(i\,\diag(\gamma_v))$.
\end{itemize}
Because $\Exp$ of a skew-symmetric matrix has determinant $+1$, the $\eta_v=\xi_v=0$ setting lies in $\SO(d)$.

\subsection{Tangent-map approximations}
The template component $\Gamma:\mathfrak g\to\mathfrak g$ is implemented in subgroup coordinates.
Let $n_{\mathfrak g}=\dim(\mathfrak g)$ and fix
\(
\mathrm{vec}_{\mathfrak g}:\mathfrak g\to\R^{n_{\mathfrak g}}
\)
with inverse
\(
\mathrm{vec}_{\mathfrak g}^{-1}
\).
For $A\in\mathfrak g$, write $a=\mathrm{vec}_{\mathfrak g}(A)$ and set
\[
\Gamma(A)=\mathrm{vec}_{\mathfrak g}^{-1}(\phi(a)).
\]
Choices used in this manuscript are:
\begin{enumerate}[leftmargin=1.5em]
  \item \textbf{Identity:} $\phi(a)=a$.
  \item \textbf{Per-direction scaling:} $\phi(a)=s\odot a$, with $s\in\R^{n_{\mathfrak g}}$.
  \item \textbf{Linear mixing:} $\phi(a)=Wa$, with $W\in\R^{n_{\mathfrak g}\times n_{\mathfrak g}}$.
\end{enumerate}
This construction is subgroup-agnostic once $\mathfrak g$ coordinates are fixed.

\subsection{Update-map approximations}
The reference group step is
\[
H\leftarrow H\Exp(A),\qquad A\in\mathfrak g.
\]
Computational approximations can replace $\Exp$ (for example, Cayley-type maps or truncated series), optionally followed by subgroup re-projection if exact subgroup constraints are required at each step.

The experiments section instantiates these choices for the reported $\Orth(d)$ runs.

\end{document}